\def\year{2021}\relax
\documentclass[letterpaper]{article} 
\usepackage{aaai21}  
\usepackage{times}  
\usepackage{helvet} 
\usepackage{courier}  
\usepackage[hyphens]{url}  
\usepackage{graphicx} 
\usepackage{csquotes}
\usepackage{natbib}  
\usepackage{caption} 
\usepackage{mwe}
\usepackage{balance}

\usepackage{amsmath}

\title{Learning Invariant Representation for Continual Learning}
\author {
    Ghada Sokar \textsuperscript{\rm 1},
    Decebal Constantin Mocanu \textsuperscript{\rm 1, \rm 2},
    Mykola Pechenizkiy \textsuperscript{\rm 1} \\
}
\affiliations {
    \textsuperscript{\rm 1} Department of Mathematics and Computer Science, Eindhoven University of Technology, The Netherlands \\
    \textsuperscript{\rm 2} Faculty of Electrical Engineering, Mathematics, and Computer Science, University of Twente, The Netherlands \\
    g.a.z.n.sokar@tue.nl, d.c.mocanu@utwente.nl, m.pechenizkiy@tue.nl
}
\begin{document}

\maketitle

\begin{abstract}
Continual learning aims to provide intelligent agents that are capable of learning continually a sequence of tasks, building on previously learned knowledge.
A key challenge in this learning paradigm is catastrophically forgetting previously learned tasks when the agent faces a new one. Current rehearsal-based methods show their success in mitigating the catastrophic forgetting problem by replaying samples from previous tasks during learning a new one. However, these methods are infeasible when the data of previous tasks is not accessible. In this work, we propose a new pseudo-rehearsal-based method, named learning Invariant Representation for Continual Learning (IRCL) \footnote{Our code is available at\\ https://github.com/GhadaSokar/Invariant-Representation-for-Continual-Learning}, in which class-invariant representation is disentangled from a conditional generative model and jointly used with class-specific representation to learn the sequential tasks. Disentangling the shared invariant representation helps to learn continually a sequence of tasks, while being more robust to forgetting and having better knowledge transfer. We focus on class incremental learning where there is no knowledge about task identity during inference. We empirically evaluate our proposed method on two well-known benchmarks for continual learning: split MNIST and split Fashion MNIST. The experimental results show that our proposed method outperforms regularization-based methods by a big margin and is better than the state-of-the-art pseudo-rehearsal-based method. Finally, we analyze the role of the shared invariant representation in mitigating the forgetting problem especially when the number of replayed samples for each previous task is small.  
\end{abstract}

\section{Introduction}
 The classic machine learning paradigm typically assumes that the data is independently and identically distributed (i.i.d.). However, this static learning paradigm contradicts the dynamic world environment which changes very rapidly. Continual lifelong learning aims to address this shortcoming, providing intelligent agents capable of adapting to the dynamic real-world environment. This capability is crucial for many systems like recommendation systems, robots, and autonomous driving where they interact with dynamic environments and operate on non-stationary data. In particular, continual learning (CL) studies the paradigm of learning a sequence of tasks, accumulating the acquired knowledge, and using it to leverage future learning.

The main challenge for applying continual learning in deep neural networks is the forgetting of previous tasks when new ones are learned, a phenomenon known as catastrophic forgetting \cite{mccloskey1989catastrophic}. This is caused by the neural networks' non-optimal ability to learn non-stationary distributions. When the model faces a new task, the weights are optimized to meet the objective of the current training samples. Therefore, the performance on previous tasks drops significantly. 

Several approaches have been proposed to address the catastrophic forgetting challenge. Regularization-based methods use a fixed-capacity model and constrain the change in important weights of previous tasks when new ones are learned. Architectural-based methods reduce the interference between tasks at the expense of expanding the model capacity. Moreover, most of these methods require the task identity during inference to select the parts of the model corresponding to the input task. This information may not be available in real-world situations. Rehearsal-based methods address the forgetting problem by replaying stored samples of previous tasks with the data of the current task. These methods have been shown to be robust against forgetting, achieving the best performance in many continual learning benchmarks \cite{VanDeVen2018a,Hsu18_EvalCL}. However, the approaches rely on replaying real data are infeasible if the data of previous tasks is inaccessible. An alternative rehearsal approach proposed by \cite{mocanu2016online, shin2017continual} is to mimic the past data using deep generative models, known as pseudo-rehearsal. Generative models have succeeded in generating high-quality images \cite{bao2017cvae,miyato2018cgans,guo2019auto}, making it possible to model the real data distribution. Another line of CL approaches is recently proposed by \cite{ebrahimi2020adversarial}. They proposed to disentangle the specific and shared representations for continual learning. They showed that the shared representation is less prone to forgetting and is effective for the continual learning paradigm. They learn the shared representation using adversarial training and use a separate specific module for each task. However, they rely on the real samples of previous tasks to mitigate forgetting. Moreover, the task identity is required during inference to pick the corresponding specific module.

\begin{figure}
\centering
\includegraphics[width=\columnwidth]{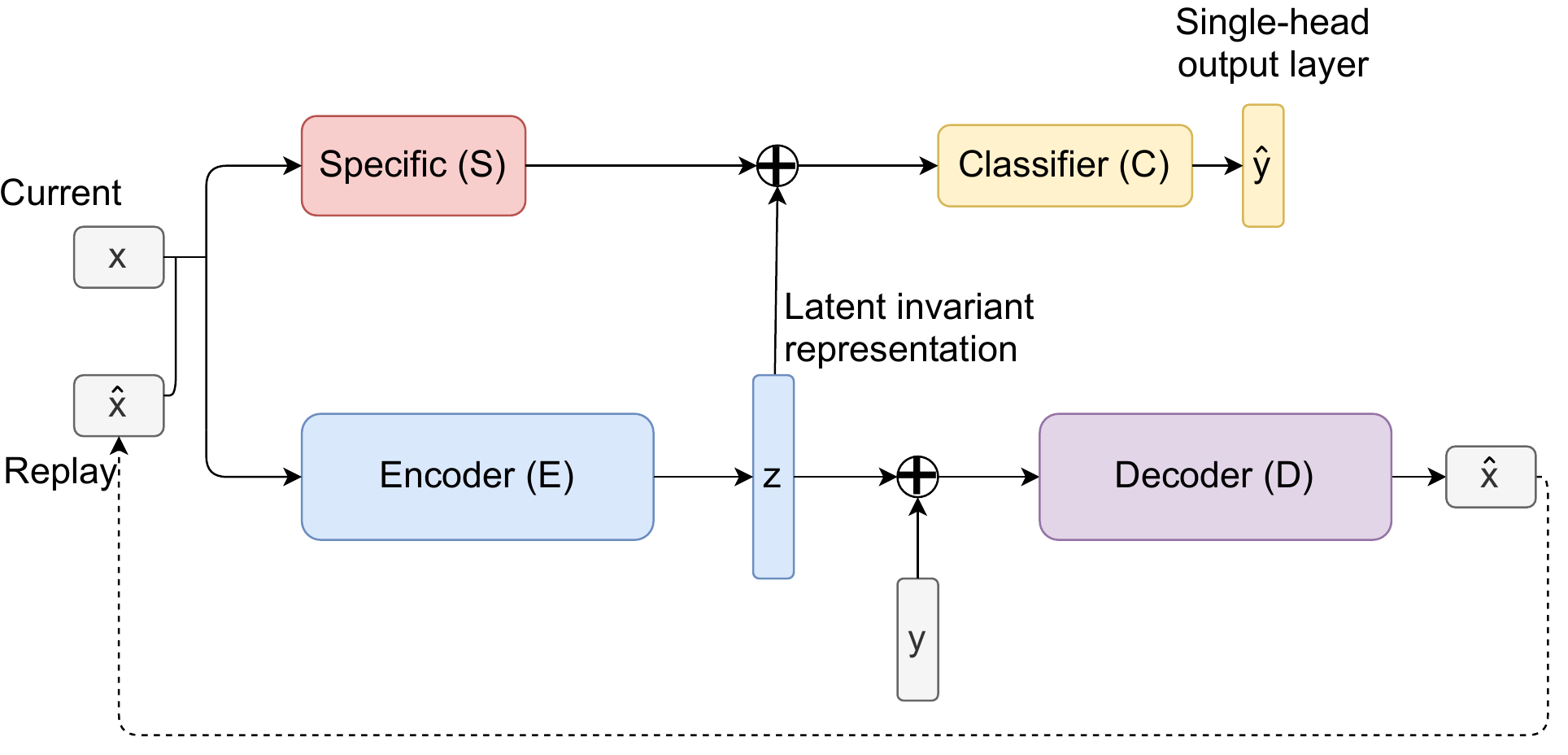}
\caption{An overview of our proposed IRCL method. Class-invariant shared representation ($z$) is learned by a conditional variational autoencoder. A specific module is used to capture discriminative representation. The two representations are jointly used for learning continually a sequence of tasks. Pseudo-samples of previous tasks ($\hat{x}$) are replayed with real data ($x$) of the current task to preserve the previously learned knowledge.}
\label{overview}
\end{figure}

In this work, we propose a new \textit{pseudo-rehearsal} approach, IRCL, for learning Invariant Representation for Continual Learning. The core idea of IRCL is to harness a conditional generative model in disentangling \textit{invariant} representation from the data which is less vulnerable to forgetting. In particular, we use a unified network for recognition and data generation as shown in Figure \ref{overview}. We factorize the space representation to an invariant representation and a specific (discriminative) representation. A conditional generative module is used for disentangling the invariant representation from the data as well as generating pseudo-samples for replay. The invariant representation is jointly used with the specific representation to perform the recognition task. The proposed approach addresses the previously stated limitations of \cite{ebrahimi2020adversarial} by using one module for capturing the specific representation and utilizing the latent shared representation for generating pseudo-samples of previous tasks instead of storing the real samples.

The contributions of this paper are: First, we propose a new method, namely IRCL, for continual learning. It learns a shared invariant representation and uses it jointly with the specific representation to continually learn a sequence of tasks while being more robust to forgetting. Second, we address the class incremental learning scenario in which the task identity is assumed to be unavailable during inference. Third, we address the cases where the real samples of previous tasks are not available. Finally, we achieve better performance than the state-of-the-art pseudo-rehearsal method by around 3.5\% and 5.5\% in terms of accuracy and backward transfer respectively on the two studied benchmarks. Moreover, the proposed method outperforms the studied regularization-based methods by a big margin.

\section{Related Work}
Continual learning has received a lot of attention from the community during the past few years. However, the idea of lifelong learning dates back to the 1990s \cite{thrun1998lifelong,thrun1995lifelong}. The existing continual learning methods can be categorized into three main strategies: regularization strategy, rehearsal strategy, and architectural strategy.

\textbf{Regularization-based methods} use fixed-capacity models and constrain the change in previous tasks by adding a regularization term to the loss function. This regularization term penalizes the change in the important weights. The weight importance is estimated by different metrics in the previous works. In Elastic Weight Consolidation (EWC) \cite{kirkpatrick2017overcoming}, an approximation of the diagonal of Fisher information matrix is used. In Synaptic Intelligence (SI) \cite{zenke2017continual}, the importance is estimated by the amount of change by weights in the loss summed over the training trajectory. The importance is computed in an online manner during training. On the other hand, in Memory Aware Synapses (MAS) \cite{aljundi2018memory}, the weights are estimated using the sensitivity of the learned function rather than the loss. Learning Without Forgetting (LWF) \cite{li2017learning} is another regularization approach that constrains the change of model predictions on previous tasks, rather than the weights, by using a distillation loss \cite{hinton2015distilling}. These approaches achieve a good performance in the task incremental learning scenario where the task identity is assumed to be known during inference. However, previous studies show their performance drop in the class incremental learning scenario \cite{kemker2018measuring,Hsu18_EvalCL,Farquhar2019a,VanDeVen2018a}.

\textbf{Rehearsal-based methods} alleviate the catastrophic forgetting by replaying samples of previous tasks. iCaRL \cite{rebuffi2017icarl}, MER \cite{riemer2018learning}, and GEM \cite{lopez2017gradient} use a small buffer to store samples from the data of previous tasks. Another direction, named generative replay, is to use a separate generative model to mimic the data of previous tasks \cite{shin2017continual,mocanu2016online}. The generated data of previous tasks is replayed along with the data of the current task to train both the generative and the classification models. Methods based on generative replay is usually called pseudo-rehearsal. 

\textbf{Architectural-based methods} alter the architecture of the network to reduce interference between tasks. Progressive Neural
Nets (PNN) instantiate a new model for each task. The previous knowledge is leveraged in learning new tasks using lateral connections to previously frozen models. Dynamic expandable network (DEN) \cite{yoon2018lifelong} splits and duplicates the neurons when the performance of previous tasks degrades much during learning a new task. The connections for each task are labeled with time-stamps to be used for inference. Other methods using sparse neural networks have been proposed for continual learning \cite{mallya2018piggyback,mallya2018packnet,sokar2020spacenet}. These methods use a fixed-capacity model and use sparse connections for each task to reduce the interference between tasks. \\
To the best of our knowledge, the very recent work by \cite{ebrahimi2020adversarial} is the only one that addresses disentangling the specific and shared representations for continual learning. They use a separate specific module for each task and the shared representation is learned using adversarial learning. During inference, one specific module is selected based on the task identity and is used with the shared representation for classifying the input data.
Most of the architectural-based methods require the task identity during inference to select the model or connections that should be used for the input. Hence, these methods are inadequate in class incremental learning.

\section{Proposed Method}
Before going into the details of our proposed method, we define the problem formulation for continual learning and the considered learning scenario. A continual learning problem consists of a sequence of tasks \{$1, 2,..., t,..., T$\}; where $T$ is the total number of tasks. Each task $t$ has its own dataset $D^{t}$. The neural network model faces tasks one by one. All samples from the current task are observed before switching to the next task. We assume that once the training of the current task ends, its data becomes not available. We focus on the class incremental learning scenario, which is more challenging yet crucial to many real-world applications like object recognition systems. In this scenario, a single-headed output layer is used for all tasks, and for each newly added task, the shared output layer is extended with the new classes in that task. Moreover, there is no knowledge about the task identity during inference. The goal is to learn the model $f_{\theta}$ that maps any input $x$ to the corresponding target class $\hat{y}$ among the classes learned so far. 

An overview of our proposed IRCL method is illustrated in Figure \ref{overview}. We have a unified network for recognition and data generation. The space representation consists of shared invariant representation learned from conditional variational autoencoder (cVAE) and specific representation to capture the class discriminative features. The motivation for disentangling the invariant representation is twofold: First, this representation is less prone to forgetting which helps in mitigating the negative backward transfer. Second, it helps in extracting the knowledge from previous tasks that could be helpful in learning future tasks. To mitigate forgetting in the shared and specific representations, pseudo-samples are generated from learned distribution by the cVAE and replayed with the real data of the current task. 

Specifically, our model consists of an encoder (E) with variational parameters $\theta_{E}$, a conditional decoder (D) with parameters $\theta_{D}$, small specific module (S) to capture discriminative representation parameterized by $\theta_{S}$, and classifier (C) with parameters $\theta_{C}$. The encoder maps an input $x$ to a latent representation $z$. The latent space is regularized by imposing a prior distribution, in our experiments a normal
distribution. The decoder is responsible for mapping the latent representation $z$ combined with the class label $y$ back to the image space $\hat{x} \sim p(x|z,y)$. Formally, the objective function of the conditional variational autoencoder (cVAE) is defined as follows:

\begin{equation}
 \mathcal{L}_{cvae} = \|x- \hat{x}\|^{2} + KL(q(z|x)\|p(z)),
\label{vaeLoss}
\end{equation}

where the first term penalizes for the reconstruction error between the original image $x$ and the reconstructed one $\hat{x}$ and the second term represents Kullback-Leibler divergence which penalizes the deviation of latent representation from the prior distribution  $z \sim \mathcal{N}(0,1)$.

Conditional generative models prove their success in generating high-quality images \cite{miyato2018cgans,odena2017conditional}. However, the main motivation for using the conditional generation is to disentangle the class-invariant representation. Previous works \cite{makhzani2015adversarial,chen2020cyclically} showed that conditioning the label information in the image reconstruction by combining it to the latent representation encourages the encoder to reduce the specific factors and retains all information independent of the label in the latent representation $z$. Therefore, the latent representation $z$ contains the class-invariant representation. In contrast to previous pseudo-rehearsal methods, where a separate model is used for the generation, we harness the latent representation $z$ as a shared representation for learning the recognition task. Since this representation is less prone to forgetting which is a crucial thing for the continual learning paradigm, we show that it helps to learn a sequence of tasks efficiently. 

However, the shared representation, being invariant, can not be used alone for performing the classification task. To this end, a small specific module (S) parameterized by $\theta_{S}$ is used to capture the discriminative representation of the input. The shared and the specific representation are jointly passed to the classifier module (C) to learn a sequence of tasks with the following objective:
\begin{equation}
 \min_{\theta_{S},\theta_{C}} \mathcal{L}_{c} (\theta_{S},\theta_{C};z, D^{t} \cup \mathcal{M}^{1:t-1}),
\label{classification_loss}
\end{equation}
where $\mathcal{L}_{c}$ is the classification loss, $D^{t}$ is the training data of the current task $t$, and $\mathcal{M}^{1:t-1}$ is the generated samples for previous tasks from $1$ to $t-1$. 

The generated samples from previous tasks are used to maintain the learned representations in the specific and shared spaces. In particular, the conditional variational autoencoder is used to draw samples from the learned distribution conditioned on the class labels of previous tasks. In contrast to previous pseudo-rehearsal methods where labels of generated images are estimated from the response of the previous version of the classification network, the conditional generation allows us to use the true labels and avoid relying on the performance of the learned classifier. The generated samples are replayed with the real samples of the current task during training the whole network. 

To train our unified network, we update the encoder, decoder, specific, and classifier modules with the following gradients respectively:

\begin{equation}
\label{gradient}
\begin{split}
  \theta_{E} \xleftarrow{} - \nabla \mathcal{L}_{cvae}, \quad
  \theta_{D} \xleftarrow{} - \nabla \mathcal{L}_{cvae},\\
  \theta_{S} \xleftarrow{} - \nabla \mathcal{L}_{c}, \quad \quad
  \theta_{C} \xleftarrow{} - \nabla \mathcal{L}_{c}.
\end{split}
\end{equation}

\section{Experiments and Results}
In this section, we evaluate the performance of our proposed method and compare it with state-of-the-art approaches for continual learning. We also analyze the representation learned by the encoder module and provide a discussion on the role of this representation in performance. 
\subsection{Benchmarks and Baselines}
We evaluate our approach on two well-known benchmarks for continual learning: split MNIST \cite{lecun1998mnist,zenke2017continual} and split Fashion MNIST \cite{xiao2017fashion,Farquhar2019a}. The MNIST dataset \cite{lecun1998mnist} consists of 70,000 images of handwritten digits from
0 to 9. The split MNIST benchmark for continual learning is introduced by \cite{zenke2017continual}. It consists of 5 tasks, each task contains two consecutive MNIST-digits. The standard training/test-split for MNIST is used resulting in 60,000 training images and 10,000 test images. The Fashion MNIST dataset is more complex than MNIST. However, it has the same size and structure of training and test sets as MNIST. The images show individual articles of clothing. This dataset is used by \cite{Farquhar2019a} to construct the split Fashion MNIST benchmark for continual learning. The benchmark consists of 5 tasks, each task has two consecutive classes of Fashion MNIST dataset. 

\textbf{Baselines}. We compared our proposed method to state-of-the-art approaches in regularization strategy. We also compare to our counterpart pseudo-rehearsal method DGR \cite{shin2017continual}. Moreover, we add another baseline in which we replay the real data of previous tasks in our proposed method, using the same splitting of the shared and specific representations, instead of replaying the generated one. This baseline provides an upper-bound to the performance of our proposed approach. We named this baseline as IRCL\_real.

\subsection{Experimental Setup}
We use a multi-layer perceptron network (MLP). Each of the encoder and decoder modules consists of one hidden layer of 300 neurons with ReLU activation. The size of the latent representation ($z$) is 32. The specific module consists of one hidden layer of a small number of neurons that equals 20. The classifier consists of one layer of 40 neurons with ReLU activation. One shared output layer is used for all tasks. Each task is trained for 5 and 10 epochs for split MNIST and split Fashion MNIST respectively. We use a batch size of 128.  The network is trained using Adam optimizer; with a learning rate of 0.01 for the cVAE and a learning rate of 0.001 for the specific module and classifier. 5000 pseudo-samples are generated for each previous task by drawing random latent representation $z$ from the normal distribution $z \sim \mathcal{N}(0,1)$ conditioned with the label information. The generated data of previous tasks is replayed with the data of the current task. 

The structure of our setup is different from the standard single network used by the baselines. However, the total number of parameters is close to the single network used by regularization methods and less than the number of parameters used by the pseudo-rehearsal baseline. For the regularization-based baselines, we used the official implementation provided by \cite{VanDeVen2018a} to evaluate their performance. The MLP network used by the baselines consists of two hidden layers, each layer has 400 neurons. For a fair comparison with the DGR method, the size of the latent representation and the number of hidden layers used for its generative model are the same as the ones used for our cVAE. More details about the experimental setup used for the baselines can be found in \cite{VanDeVen2018a}.

\subsection{Evaluation Metrics}
We evaluate the performance of our method using two evaluation metrics. The first one is the average classification accuracy (ACC) across all tasks after training the whole sequence. This metric is commonly used in the literature for continual learning. The second one is the backward transfer metric (BWT) which is proposed in \cite{lopez2017gradient}. This metric indicates how much learning a new task has influenced the performance of previous tasks. Higher negative values for BWT indicate catastrophic forgetting. Formally, ACC and BWT are calculated as follows: 
\begin{equation}
\label{ACC_BWT}
\begin{split}
   ACC = \frac{1}{T} \sum_{i=1}^{T} R_{T,i}, \\
   BWT = \frac{1}{T-1} \sum_{i=1}^{T-1} R_{T,i} - R_{i,i}, 
\end{split}
\end{equation}
where $R_{j,i}$ is the accuracy on task ${i}$ after learning
the $j$-th task in the sequence, and $T$ is the total number of tasks.

\subsection{Results and Analysis}
\begin{figure}[hb]
\centering
\includegraphics[width=0.95\columnwidth]{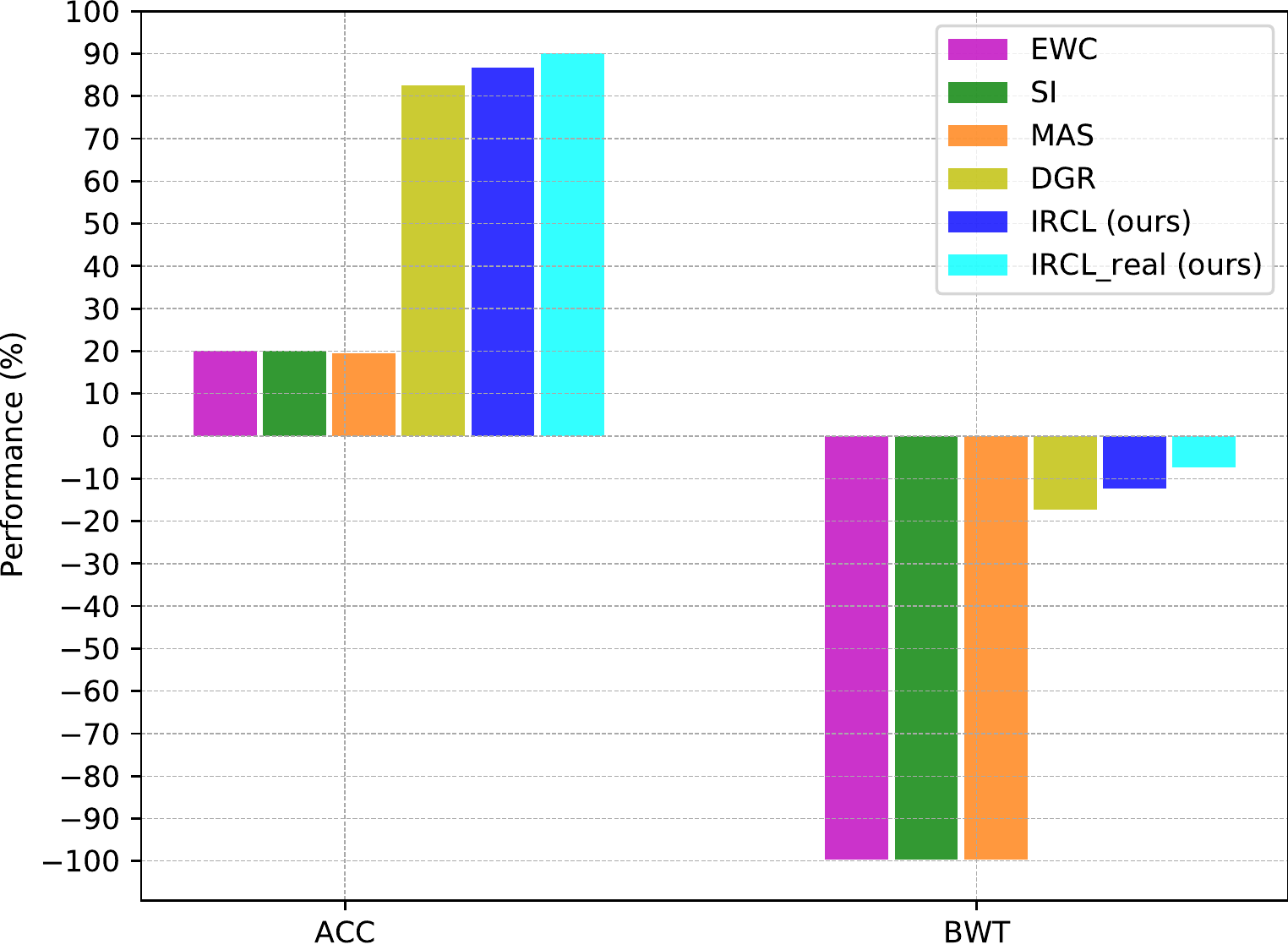}
\caption{ACC and BWT on the split MNIST benchmark.}
\label{splitMNIST}
\end{figure}

\begin{figure}[ht]
\centering
\includegraphics[width=0.95\columnwidth]{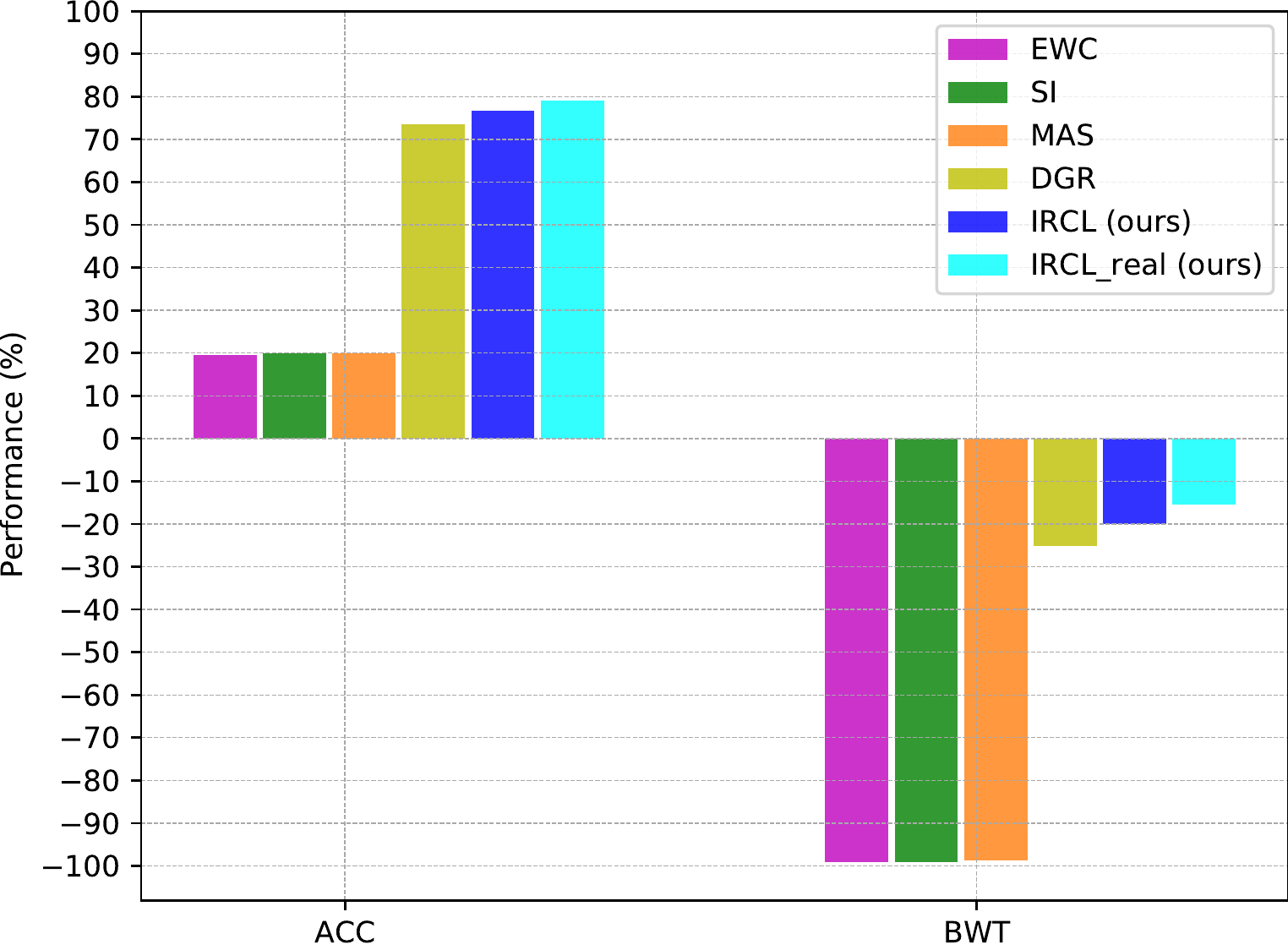}
\caption{ACC and BWT on the split Fashion MNIST benchmark.}
\label{splitFashionMNIST}
\end{figure}

Figure \ref{splitMNIST} and Figure \ref{splitFashionMNIST} show the ACC and BWT on the split MNIST and split Fashion MNIST benchmarks respectively. Consistent with the finding of previous studies, the regularization methods catastrophically forget the previously learned tasks. They only have a high performance for the last task causing a very high negative backward transfer. The pseudo-rehearsal method, DGR, manages to maintain the performance of previous tasks. Our proposed method achieves a better performance than the DGR method on the two benchmarks while using a unified network with a smaller number of parameters. On the split MNIST  
 benchmark, our proposed method outperforms the DGR method in terms of ACC by 4\%. It also achieves lower negative backward transfer; 5\% less than the DGR method. On the split Fashion MNIST benchmark, our proposed method achieves higher ACC than DGR by 3\% while achieving 6\% less in terms of negative BWT. It is also worth to be highlighted that the proposed method achieves a very close performance \enquote{IRCL\_real} baseline in which we replay the real data of previous tasks rather than replaying the generated ones.\\

\begin{figure}[ht]
\centering
\includegraphics[width=\columnwidth]{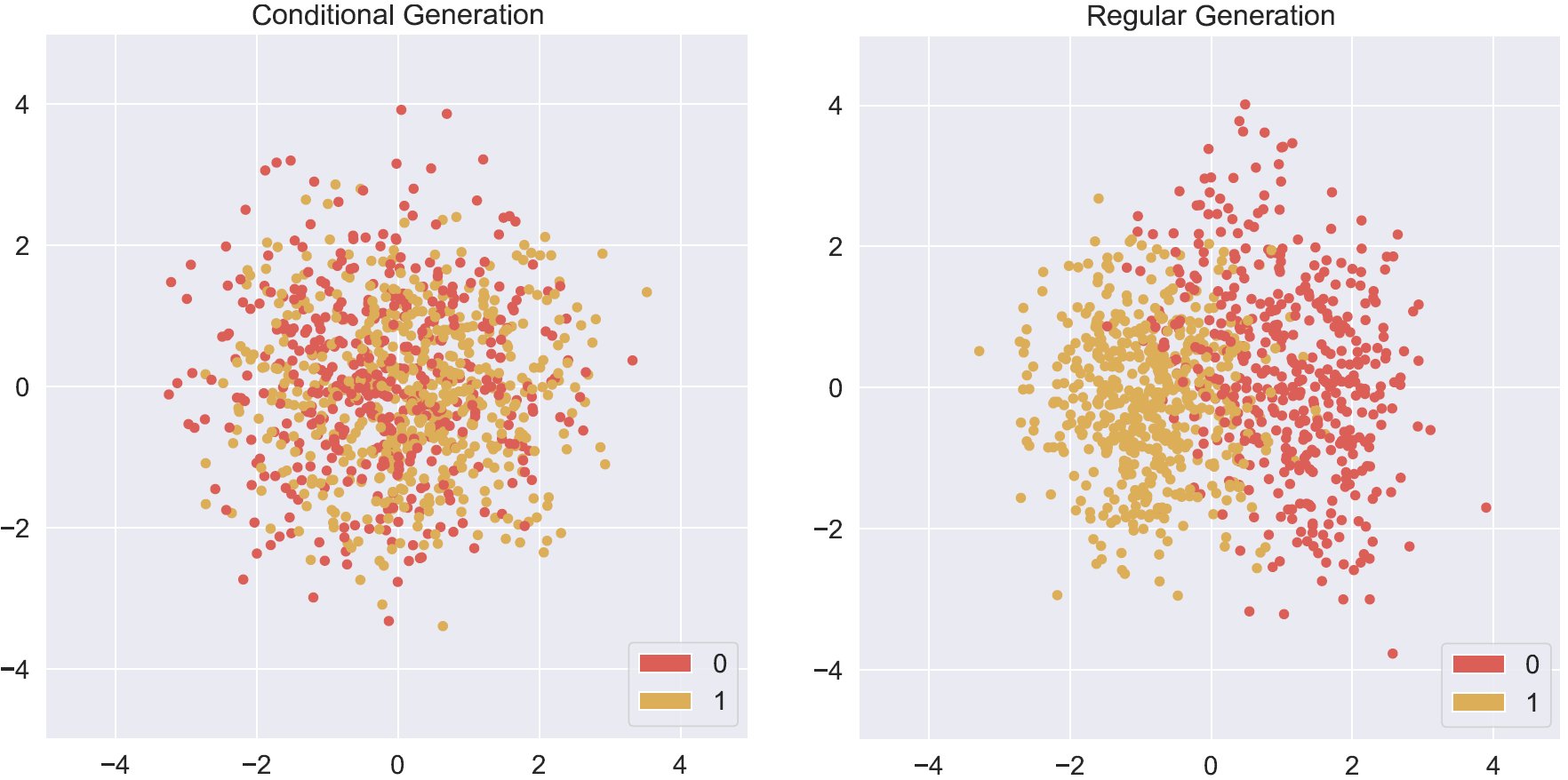}
\caption{Visualizing the effect of conditioning the label information on the latent representation using random samples from the test data of the first task (classes 0,1) of split MNIST.}
\label{z_with_conditioning}
\end{figure}

\begin{figure}[ht]
\centering
\includegraphics[width=\columnwidth]{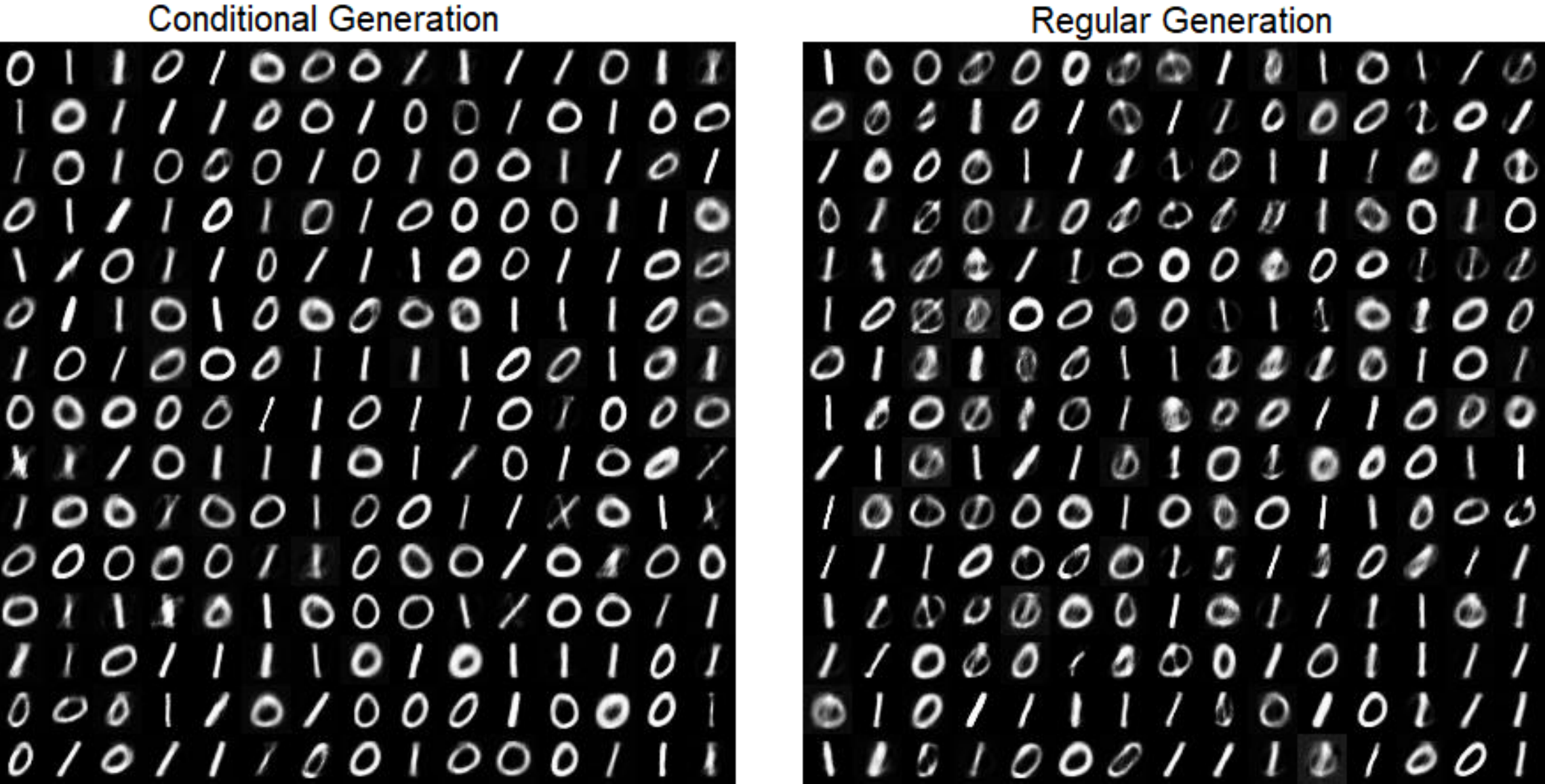}
\caption{Visualizing the effect of conditioning the label information on the generated samples of the first task (classes 0,1) of split MNIST.}
\label{samples_with_without_condition}
\end{figure}

\textbf{Conditional generation} Next, we analyze the role of conditional generation by feeding the class labels into the decoder module in disentangling the shared representation. We visualize the latent representation $z$ using Principal Component Analysis (PCA) \cite{wold1987principal} in two cases: (1) conditioning the label information during generation and (2) the regular generation without conditioning. We performed this analysis on the split MNIST benchmark. We draw the latent representation of 1000 random samples from the test data of the first task. As shown in Figure \ref{z_with_conditioning}, the latent representation learned in the case of conditional generation is class-invariant. Providing the generative model with the label information encourages the encoder to reduce the specific factors in the latent representation. While in the other case, the representation is discriminative. We have also observed that the conditional generation has another effect on the generated image quality. Figure \ref{samples_with_without_condition} shows samples of the generated data from conditional and regular generative models. The figure illustrates that the quality of the generated data from the conditional generative model is better than the regular generative one. Having good quality images for previous tasks also helps in maintaining their performance.  Finally, recent work by \cite{aljundi2019online} shows that learning a new task does not equally interfere with previously learned tasks. Conditional generation in our proposed method facilitates controlling the number of generated images from each class.\\

\begin{figure}[ht]
\centering
\includegraphics[width=0.8\linewidth]{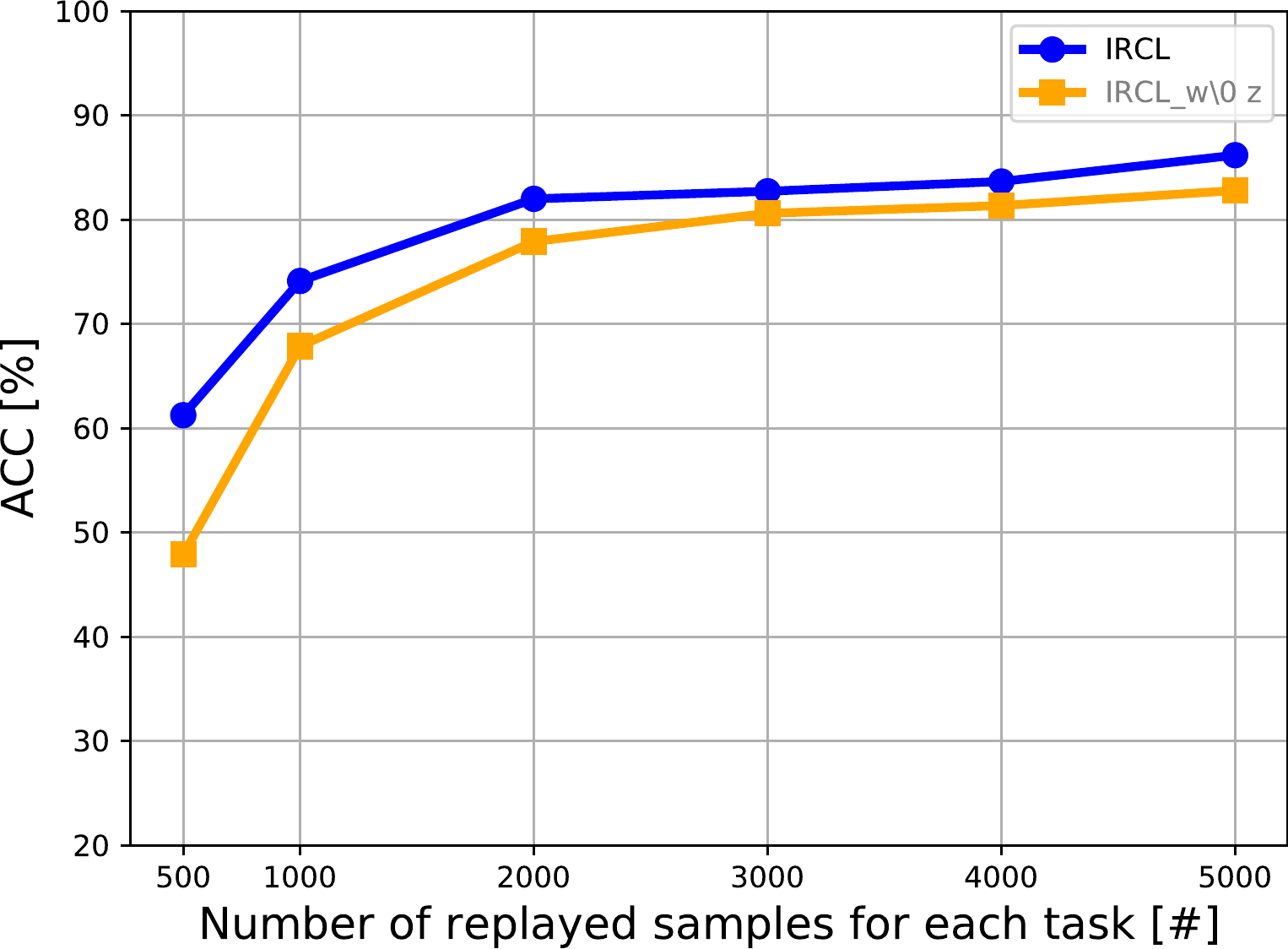}\\
\includegraphics[width=0.8\linewidth]{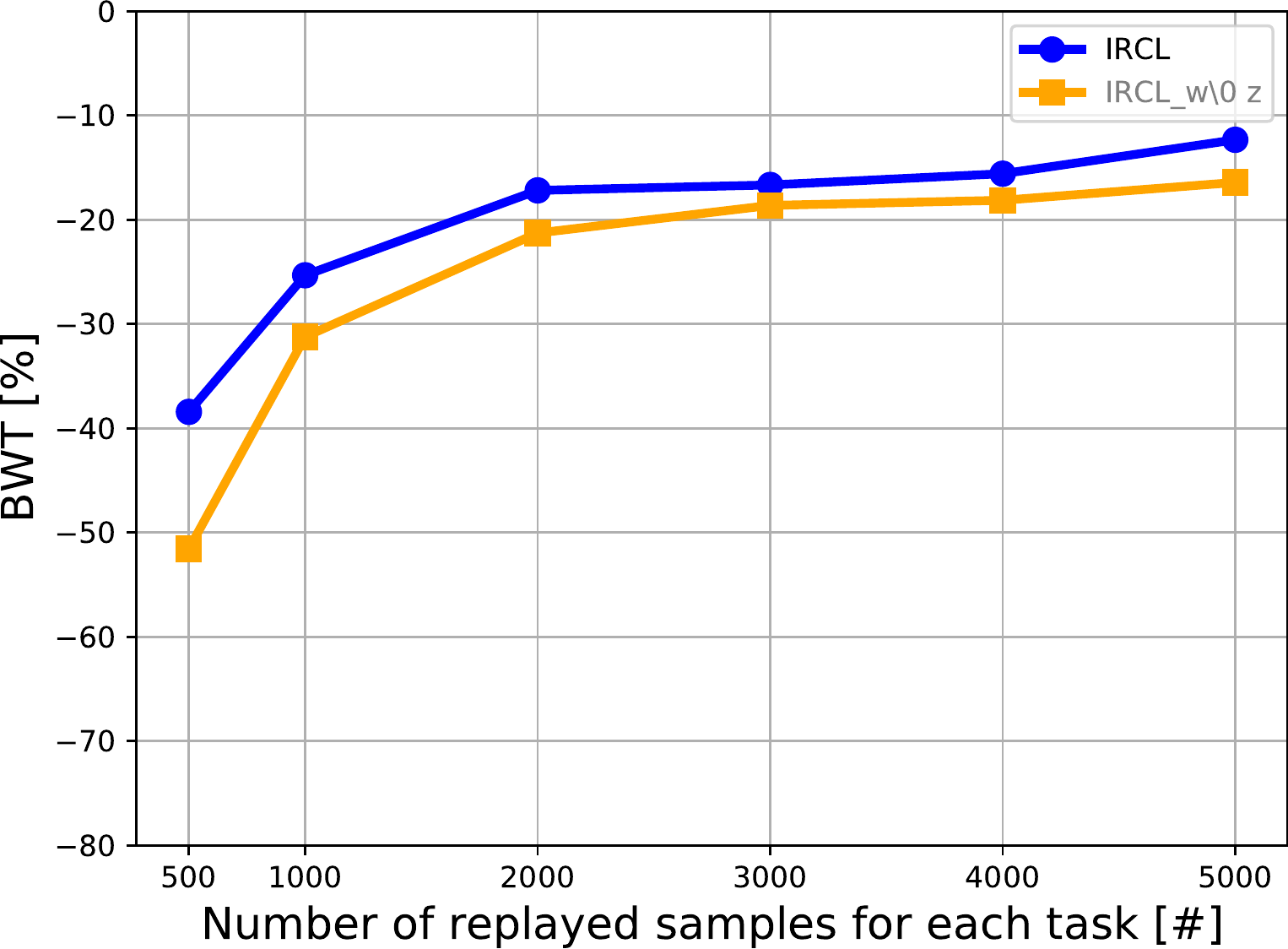}
\caption{Comparison of the effect of providing the $z$ representation to the classifier. ACC and BWT are reported on split MNIST using a different number of replayed samples for each task.}
\label{z_with_conditioning_mnist}
\end{figure}

\begin{figure}[ht]
\centering
\includegraphics[width=0.8\linewidth]{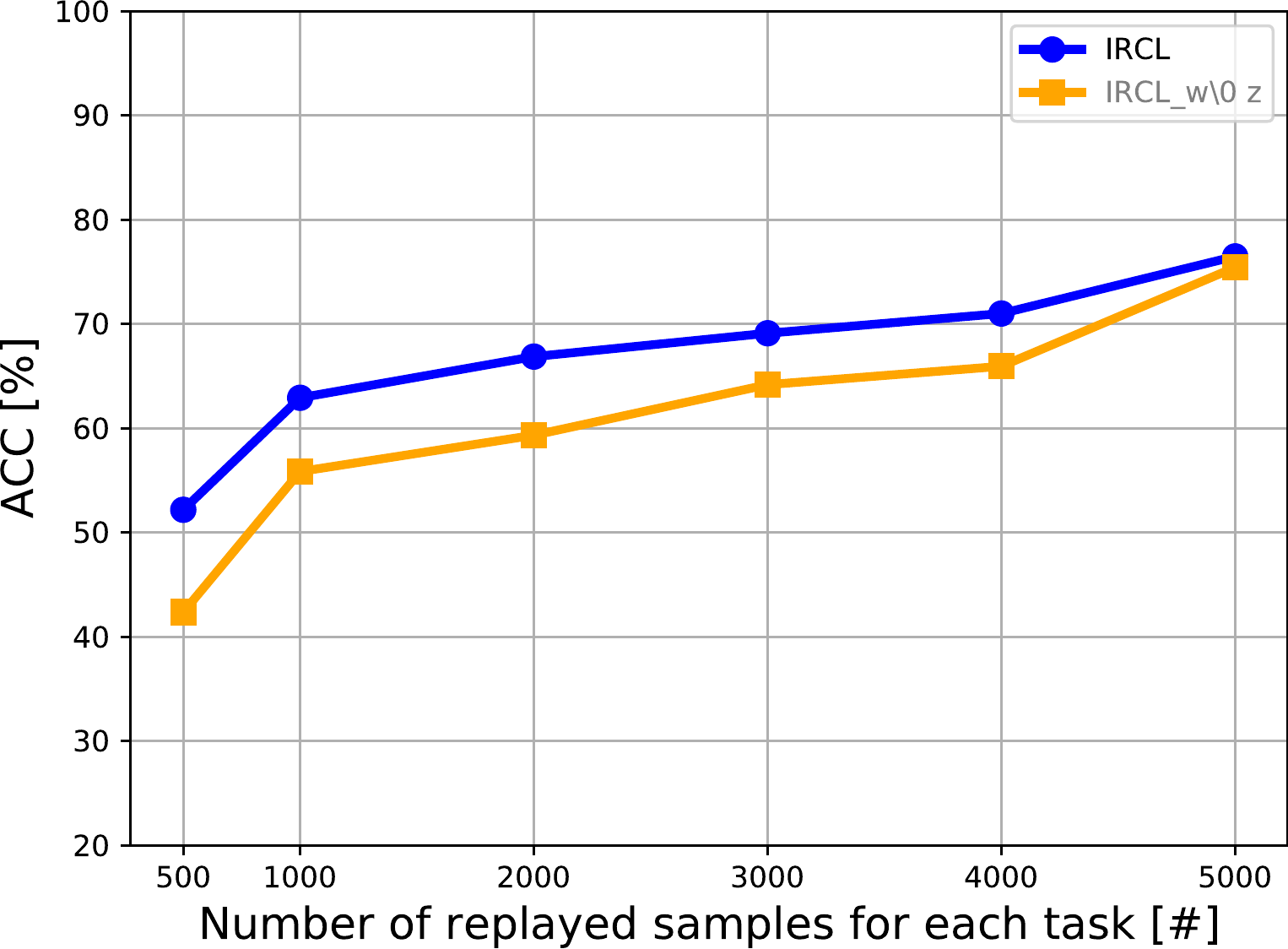}\\
\includegraphics[width=0.8\linewidth]{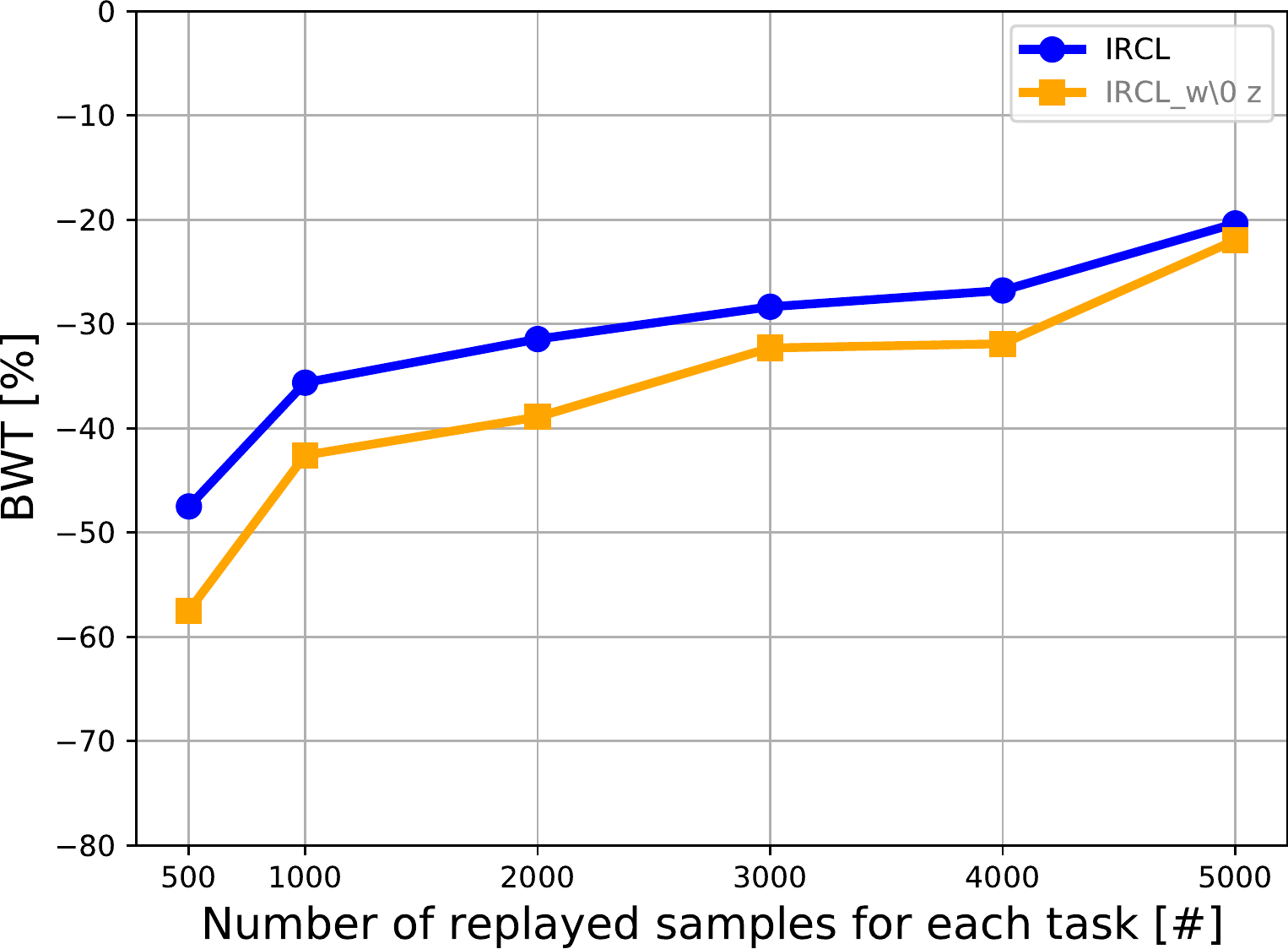}
\caption{Comparison of the effect of providing the $z$ representation to the classifier. ACC and BWT are reported on split Fashion MNIST using a different number of replayed samples for each task.}
\label{z_with_conditioning_fashion} 
\end{figure}

\textbf{Shared representation} Here we analyze the role of feeding the shared invariant representation to the classifier. We held this experiment on a varying number of replayed samples to study its effect in cases where the model is more prone to forgetting. In particular, we held a comparison between two cases: (1) the original proposed IRCL method in which the latent representation $z$ is provided to the classifier and (2) \enquote{IRCL\_w\textbackslash o $z$} baseline in which the shared representation is not provided to the classifier. We calculate the ACC and BWT on the two studied benchmarks using a different number of replayed samples. Figure \ref{z_with_conditioning_mnist} and Figure \ref{z_with_conditioning_fashion} show the results on split MNIST and split Fashion MNIST respectively. Providing shared representation to the classification module always has a better performance. The figures also show that providing shared representation helps in reducing the negative backward transfer. The importance of this representation increases when the number of replayed samples is small and the model is more prone to forgetting. This reveals that the shared representation is less prone to forgetting and can help in maintaining the performance of previous tasks.  

\section{Conclusion}
In this paper, we have proposed IRCL, a new simple yet effective method to provide deep neural networks the ability to learn a sequence of tasks continually. IRCL learns a shared invariant representation using a conditional variational autoencoder (cVAE). This representation is jointly used with specific representation for classification. Pseudo-rehearsal is used to maintain previously learned knowledge. We consider the class incremental learning scenario in which the task identity is not available during inference. The proposed method is evaluated on two standard benchmarks for CL: split MNIST and split Fashion MNIST. Experimental results show the effectiveness of our method and its robustness towards forgetting. Our proposed method outperforms the regularization methods by a large margin. It also achieves a better performance than the state-of-the-art pseudo-rehearsal method by 4\% and 3\% on split MNIST and split Fashion MNIST respectively. It also reduces the negative backward transfer by 5\%  and 6\% on the two benchmarks respectively. 
Moreover, we show that shared learned representation by the cVAE helps in mitigate forgetting especially when the number of replayed samples from previous tasks is small. 

In the future, it would be interesting to analyze the effect of adding more constraints to guarantee that the shared representation does not appear in the class-specific representation. Another interesting extension to the work would be to generate a different number of samples for each task based on how much learning a new task affects its performance. The conditional generation module of our proposed method gives the facility to control the number of generated images for a specific task.

\balance
\bibliography{mybibfile}

\end{document}